\documentclass[10pt, a4paper]{article}
\usepackage{lrec}
\usepackage{graphicx}
\usepackage{tabularx}
\usepackage{soul}
\usepackage{epstopdf} 
\usepackage[utf8]{inputenc}
\usepackage[T1]{fontenc}
\usepackage[unicode]{hyperref}
\usepackage[noabbrev,capitalize]{cleveref}
\usepackage{xstring}

\def\ptakopet{Ptakopět}

\title{Outbound Translation User Interface Ptakopět: A Pilot Study}

\name{Vilém Zouhar, Ondřej Bojar}

\address{Charles University \\
         Prague, Czech Republic \\
         \{zouhar, bojar\}@ufal.mff.cuni.cz\\}
    
\usepackage{color}
\usepackage{multirow}
\usepackage{enumitem}
\usepackage{float} 
\setlist[itemize]{itemsep=-2pt,leftmargin=*}
\newcommand{\itemname}[1]{\textbf{#1}}
\newcommand{\footnotehref}[2]{%
\footnote{\href{#1}{#2}}%
}

\def\perscite#1{\newcite{#1}} 

\abstract{
It is not uncommon for Internet users to have to produce a text in a foreign language they have very little knowledge of and are unable to verify the translation quality. We call the task ``outbound translation'' and explore it by introducing an open-source modular system Ptakopět. Its main purpose is to inspect human interaction with MT systems enhanced with additional subsystems, such as backward translation and quality estimation. We follow up with an experiment on (Czech) human annotators tasked to produce questions in a language they do not speak (German), with the help of Ptakopět. We focus on three real-world use cases (communication with IT support, describing administrative issues and asking encyclopedic questions) from which we gain insight into different strategies users take when faced with outbound translation tasks. Round trip translation is known to be unreliable for evaluating MT systems but our experimental evaluation documents that it works very well for users, at least on MT systems of mid-range quality. \\ \ \\ \
}

\begin{document}

\maketitleabstract

\section{Introduction}

For most language pairs, machine translation (MT) quality is limited. Nevertheless, MT in everyday use greatly helps by providing low quality, preview translation also called gisting. The complement of gisting is outbound translation. In both cases, a message is transferred between the author and the recipient and each of them has sufficient knowledge of only their language. In outbound translation, the author is responsible for creating correct messages in the recipient's language. In gisting, the message is sent in the author's language and the responsibility to correctly interpret it lies on the recipient. An example of gisting would be browsing on a website in a foreign language, whilst filling in a form in a foreign language would be an example of outbound translation.

When translating to foreign languages, users cooperate with machine translation tools to produce the best result. Machine translation can prepare a first version of the text, or it can be used to verify the user's own translation to some extent.

Users translating into languages which they do not master enough to validate the translation need some additional system for verification and assurance that the machine translation output is valid. For this, Ptakopět offers word-level quality estimation (QE), simulated source complexity and backward translation. While round trip translation may be unreliable for fully automatic evaluation of MT quality \cite{round_trip_translation}, it is still a widespread strategy for users. 

The paper is structured as follows: We briefly introduce the components we rely on in \cref{background} and describe \ptakopet{} in \cref{ptakopet}, including the underlying models. The experiment setup is presented in \cref{experiment} and the results in \cref{results} We conclude in \cref{conclusion}

All gathered data is stored in a public repository.\footnotehref{https://github.com/zouharvi/ptakopet}{https://github.com/zouharvi/ptakopet}

\section{Background}
\label{background}

\subsection{Quality Estimation}
\label{qe}

Machine translation quality estimation is used mostly in translation companies to minimize post-editing costs. Unfortunately, quality estimation cues are missing in most of the mainstream public translation services, such as Google Translate\footnotehref{https://translate.google.com/}{translate.google.com} (provides alternatives to words), Microsoft Translator\footnotehref{https://www.bing.com/translator}{bing.com/translator} or DeepL\footnotehref{https://www.deepl.com/en/translator}{deepl.com/en/translator} (provides alternatives to phrases).

Quality estimation is usually performed on bitext (parallel text composed of source and target language versions). The four levels with the following metrics, as distinguished by the WMT shared task \cite{wmt19qe} are:

\begin{itemize}
    \item \itemname{word-level} -- words in a target sentence are classified as OK or BAD 
    \item \itemname{phrase-level} -- phrases in a target sentence are classified as OK or BAD
    \item \itemname{sentence-level} -- target sentence receives a score, such as percentage of edits needed to be fixed: HTER, post-editing time in seconds, or counts of various types of keystrokes.
    \item \itemname{document-level} - target document gets an MQM score\footnotehref{http://www.qt21.eu/mqm-definition/definition-2015-12-30.html}{qt21.eu/mqm-definition/definition-2015-12-30.html}
\end{itemize}

For our case, only word or phrase-level quality estimates are sufficiently
informative.

\begin{figure*}
    \centering
    \includegraphics[height=\textwidth, angle=-90]{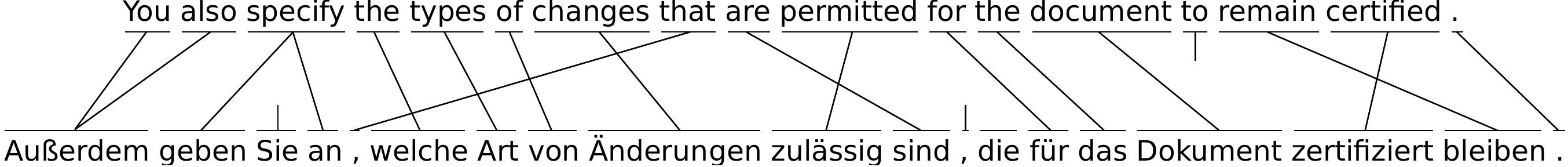}
    \caption{\label{fig:alignment} Word alignment of the first sentence in WMT17 shared task 2 training data (English to German)}
\end{figure*}

\subsection{Word Alignment}




Word alignment is the task of matching two groups of words in a sentence pair if and only if they are each other's translations. Word alignment usually follows after sentence alignment. An example of word alignment between an English sentence and translated German sentence can be seen in \cref{fig:alignment}.

Word alignment in Ptakopět is used to tell users which parts of their source sentences were probably poorly translated. In the context of outbound translation, highlighting parts of the translated sentence provides only little information to the user, since they do not know what they map to in the source sentence.

Ptakopět highlights words in the source sentence with the same intensity as the
matching words in the target sentence. The same form of assistance could also be
provided directly using some form of a source complexity estimator instead of the combination of quality estimation and word alignment.

\section{Ptakopět}
\label{ptakopet}

\ptakopet{} is a modular system implemented primarily in TypeScript (frontend) and Python (backend), interfacing external text processing components using web sockets or Unix pipes.

\cref{backfront} introduces the frontend-backend structure of \ptakopet{}. \cref{ui} illustrates the current user interface. We then describe the particular MT system chosen for our experiment (\cref{mtsys}), the quality estimation models and training data for our language pair of interest (\cref{qesys})

\subsection{Backend and Frontend}
\label{backfront}

The Ptakopět backend\footnotehref{https://github.com/zouharvi/ptakopet-server}{github.com/zouharvi/ptakopet-server} is a simple server with a queue that responds to quality estimation and word alignment requests. Apart from that, it serves as a logger endpoint for experiments. The Ptakopět frontend\footnotehref{https://github.com/zouharvi/ptakopet}{github.com/zouharvi/ptakopet} is a web page\footnotehref{https://ptakopet.vilda.net}{ptakopet.vilda.net} which allows the users to translate texts with the help of quality estimation (highlighting poorly translated words) and backward translation. It was designed so that more components can be added and different approaches tried.

Both the server and the frontend can be run and installed locally. Technical details with instructions are in the online documentation.\footnotehref{https://ptakopet.vilda.net/docs}{ptakopet.vilda.net/docs}

\begin{figure}[ht]
    \centering
    \includegraphics[width=0.99\columnwidth]{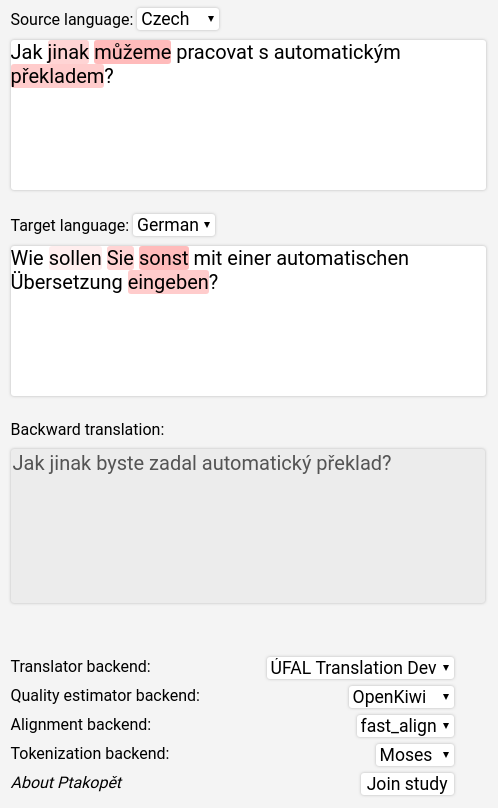}
    \caption{\label{fig:ptakopet_ui_1} Example sentence in Ptakopět with quality estimation highlighting and backward translation}
\end{figure}

\subsection{User Interface}
\label{ui}

The main Ptakopět layout is displayed in \cref{fig:ptakopet_ui_1}. It contains three text areas. The top-most is the input field for text in the source language. Underneath follows translation to the foreign language and bottom-most is the backward translation. Quality estimation is performed on the texts in the first and the second input fields (source and forward translation) and is rendered in the latter. Quality estimation is then transferred via word alignment to the source text and shown there.


\begin{figure*}[t]
    \centering
    \includegraphics[height=\textwidth, angle=-90]{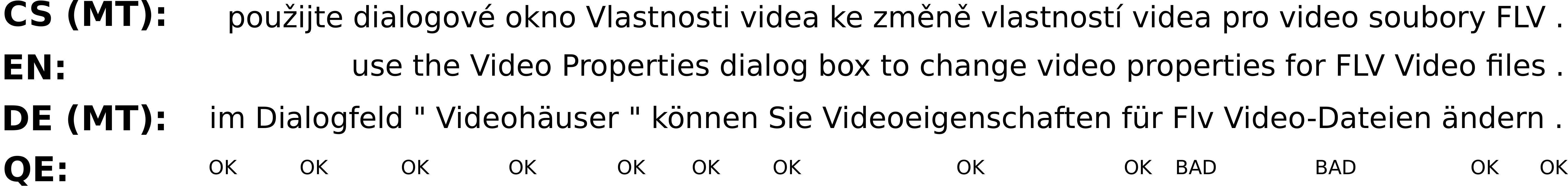}
    \caption{\label{fig:qe_example} Quality estimation tags for tokens and gaps in a German sentence translated from English (from WMT19 quality estimation shared task) together with synthetic Czech source (translated from English). MT systems are independent.}
\end{figure*}

\subsection{Machine Translation models}
\label{mtsys}

\ptakopet{} is flexible in terms of the underlying MT engine and even allows the user to choose the engine on the fly with a drop-down menu. For the purposes of our experiment, we stick to one particular engine of mid-range translation quality. We motivate the choice by the fact that very high-quality MT is available only for a handful of language pairs and these language pairs may not need any support in outbound translation.

We made use of two neural MT Transformer models (CS$\rightarrow$DE and DE$\rightarrow$CS) described in Section 7 in \cite{mt_machacek}. They were trained on 8.8M Czech-German sentence pairs for eight days from scratch until convergence. 

For performance sake, the system accepts only a limited number of subword units per translation computation. Most sentences fit into this limit, but longer sentences do not, which results in context loss. Generally, both MT models made mistakes occasionally, such as adding extra words or phrases.

\subsection{Quality Estimation models}
\label{qesys}

Ptakopět uses quality estimation for highlighting poorly translated words. There were three available implementations, but none of them was suitable for online use out of the box. We also synthesized QE training data for our language pair of interest, see \cref{subsec:cs_de_wmt}

\subsubsection{QuEst++}
The main pipeline of QuEst++ \cite{questplusplus} consists of feature extraction and machine learning prediction. It first extracts features WMT12-13-14-17\footnotehref{https://www.quest.dcs.shef.ac.uk/quest\_files/features\_blackbox\_baseline\_17}{quest.dcs.shef.ac.uk/quest\_files/features\_blackbox\_\\baseline\_17} from the input data, such as POS, indication of words' presence in a dictionary and word length and then runs a standard ML algorithm e.g. Cross-validated Lasso, using the LARS algorithm. Especially the feature extraction part is not optimized and it is quite slow.

The original feature extractor system supports English-Spanish quality estimation. We experimented with feeding it English-Czech quality estimation data and expected that the ML part would disregard noisy or low information features caused by feeding the feature extractor unsupported language. We found that the performance regressed so considerably (even on the training data) that we did not experiment further with Czech-German quality estimation in QuEst++.

\subsubsection{DeepQuest}

DeepQuest \cite{deepquest} takes a neural approach to quality estimation and is capable of performing well on any language pair. The toolkit offers two architectures: a reimplementation of Predictor-Estimator architecture \cite{Kim-Postech:2017} and a bidirectional recurrent neural network (bRNN) system. DeepQuest offers document-level, sentence-level, phrase-level and word-level quality estimation.

We trained the bRNN model on WMT17 English-German data and synthesized WMT17 Czech-German data described below. This architecture does not require pretraining, is less complex and provides results close to Predictor-Estimator \cite{deepquest}.

\subsubsection{OpenKiwi}

OpenKiwi \cite{openkiwi} implements three quality estimation models: QUality Estimation from ScraTCH \cite{kreutzer-quetch:2015}, NuQE \cite{martins-unbabel:2016} used for WMT19\footnotehref{http://www.statmt.org/wmt19/qe-task.html}{statmt.org/wmt19/qe-task.html} baseline and Predictor-Estimator \cite{Kim-Postech:2017}. Additionally, OpenKiwi implements stacked ensembling as proposed by \perscite{martins-ms:2017}.

We opted for the Predictor-Estimator architecture for our experiment, because even though it requires pretraining, it does not consume so many resources compared to the stacked ensemble. This architecture also provides the best results in comparison with other architectures without ensembling, as shown in \cite{openkiwi}.

OpenKiwi, in general, proved to be faster, more robust and easier to use than DeepQuest. Because of this, the experiment was conducted with OpenKiwi quality estimation backend.

\subsubsection{Czech-German Quality Estimation dataset} \label{subsec:cs_de_wmt}

Since relevant Czech-German training data for QE were not available, we synthesized them from WMT 2017 English-German Word Level Quality Estimation dataset in the IT domain \cite{WMT17}. Such data are composed of source language sentences (EN), target language sentences (DE) and OK/BAD tags for each word (QE).

We processed the WMT17 English-German data to obtain Czech-German data by translating the source language sentences using LINDAT Translation \cite{popel-en-cs} from English to Czech. Given triplets (EN, DE, QE), we thus create triplets of (CS, DE, QE). An example of this can be seen in \cref{fig:qe_example}.

To make sure the data did not lose quality, we performed the following experiment: We manually annotated 30 Czech-German and 20 English-German sentences for word-level quality estimation, in the same format as the original English-German dataset, i.e. labelling German words with OK/BAD labels given the source sentence. The original English-German annotation served as the golden standard. Our annotation for English-German was created independently of it and it served as a benchmark for our agreement with the original. 

\begin{table}[ht]
    \centering
    \begin{tabular}{| l l |}
        \hline
        \multicolumn{2}{|c|}{All}\\
        \hline
        TP=74.57\%& FP=2.68\%\\
        FN=12.98\%& TN=9.76\%\\
        \hline
        \multicolumn{2}{|c|}{Czech-German}\\
        \hline
        TP=77.58\%& FP=3.68\%\\
        FN=11.03\%& TN=7.71\%\\
        \hline
        \multicolumn{2}{|c|}{English-German}\\
        \hline
        TP=69.81\%& FP=1.11\%\\
        FN=16.07\%& TN=13.02\%\\
        \hline
    \end{tabular}
    \caption{\label{tab:manual_qe_annotation}Confusion matricies for word level quality estimation annotations of Czech-German and English-German. (TP = True positive, FP = False positive, TN = True negative, FN = False negative)}
\end{table}

\cref{tab:manual_qe_annotation} shows the confusion matrices of our annotations compared to the golden standard. The distributions for both language pairs are similar. The sample is very small and the sets of underlying sentences (20 English and 30 Czech) had to be different because the annotation was carried out by a single person, but the results nevertheless indicate that this transfer of QE data by machine-translating the source is viable. The similarity of confusion scores can mean one of the following. Either the German sentence itself was representative enough for the annotator to produce classes with similar distributions, or that both the English and the Czech sentences provided the same level information. In both cases, the pairs (EN, DE) and (CS, DE) seem equally usable which means that we should be able to train similarly good quality estimation model based on the synthetic Czech source.

\subsection{Alignment}

We use Hunalign \cite{varga:2007} for sentence alignment and fast align \cite{fastalign:2013} for word alignment, both because of their ease of use and good performance.

Both sentence and word alignment systems are unsupervised, operating only on the given input data. Because the real input received by \ptakopet{} is generally very short, we always mix it with a baseline parallel corpus. This increases the vocabulary coverage for word alignment and improves the stability of sentence alignment.

The training data for quality estimation (\cref{subsec:cs_de_wmt}) already limited us to the IT domain. We thus choose a similar domain also for this additional corpus for alignment, the widely available Ubuntu 14.10 parallel corpora \cite{opus_ubuntu}. Specifically, we use parallel corpora for the following language pairs: EN-CS (6492 sentence pairs), DE-CS (6604 sentence pairs), DE-EN (13245 sentence pairs), CS-FR (6603 sentence pairs), EN-FR (9375 sentence pairs).
These corpora are used both for word and sentence alignment.

\section{Experiment Setup}
\label{experiment}

The goal of our pilot experiment was to observe and describe strategies users take when tasked to do outbound translation and see if and how \ptakopet{} helps in the task.

The experiment was carried out remotely, in two phases. In the first phase, annotators were presented with a sequence of web pages in Czech or English and asked to produce a German sentence given a stimulus at each of them. In the second phase (\cref{validation}), a highly-skilled speaker of German validated the outputs of the first phase.

QE highlighting in \ptakopet{} was enabled only for IT domain stimuli, because the QE model did not perform well on out of domain sentences. 

\subsection{Annotators}

For the first phase, there were 8 Czech annotators in total, divided into two groups. The first one was composed of 4 people without advanced knowledge of English\footnote{Note that the annotators never needed to produce any English text in the experiment. Only one subset of the test data needed English comprehension.} and the second one consisted of 4 people with English level of at least C1 on the CEFR scale. All of the annotators had German knowledge of at most A1. We refer to these groups as bilingual and monolingual, respectively.

\subsection{Data}

For our experiment, we gathered input data and prompted users to reformulate a specific question or work with the text in some way. Each data section was meant to correspond to a real-life application.

\subsubsection{Seeking help in technical issues}

For the best match with the QE training data (\cref{subsec:cs_de_wmt}), we extracted 35 stimuli (in Czech) from WMT 2017 English-German quality estimation dataset. The sentences describe technical issues when using common office or desktop publishing programs.

The annotators were expected to translate the description of the issue to German relying on machine translation and quality estimation tools.
Furthermore, we think that explaining technical issues to IT support in an unknown language is a common outbound translation use case. An example of a technical issue is in \cref{fig:tech_example} (translated to English).

\begin{figure}[ht]
    \noindent\rule{\linewidth}{0.5pt}
    \textbf{Issue description:}
    
    The date format cannot be changed from Month-Day-Year to Day-Month-Year.
    
    \vspace{-0.2cm}\noindent\rule{\linewidth}{0.5pt}\vspace{-0.3cm}
    \caption{\label{fig:tech_example} Example description of a technical issue
    from the experiment dataset.}
\end{figure}

\subsubsection{Common administrative issues}

The next 30 stimuli in the experiments provided a source text in Czech with a piece of factual information (a short span in the text) highlighted. The annotators were supposed to formulate questions that ask for this factual information.

This data was collected from the instructions on how to proceed in various administrative topics at the Municipal District of Prague 6.\footnotehref{https://www.praha6.cz/codelat/index.php}{praha6.cz/codelat/index.php} This use case is inspired by the day to day problems of citizens living in a foreign city. With the help of MT, they can get the gist of a regulation or relevant document, but they may need to ask the administration for some clarification or a specific detail.


An example of an administrative issue stimulus can be seen in \cref{fig:admin_example}. For presentation purposes, we again translate the stimulus into English, but the annotators saw Czech text and were expected to formulate the question in Czech so that MT produces a good German version.

\begin{figure}[ht]
    \noindent\rule{\linewidth}{0.5pt}
    \textbf{Paragraph with span:}
    
    Applicant pays {\underline{100 CZK}} when changing a surname that is derogatory, eccentric, ridiculous, garbled or foreign.
    
    \vspace{-0.2cm}\noindent\rule{\linewidth}{0.5pt}\vspace{-0.2cm}
    \caption{\label{fig:admin_example} Example administrative topic and the
    factual information to ask for (the price) highlighted}
\end{figure}

\subsubsection{Encyclopedic knowledge: SQuAD 2.0}

The last section of the experimental data was based on the Stanford Question Answering Dataset 2.0 \cite{SQUAD-2.0} and its (machine-translated) Czech version. The basic unit of SQuAD are paragraphs with spans. In the context of SQuAD 2.0, this means that there already existed a question for this span. In our experiment, we disregard the existing questions and ask our annotators to ask for the highlighted information again. We are thus creating additional questions for the SQuAD dataset, now in Czech.

An example of a paragraph from SQuAD 2.0 and questions we collected from the Ptakopět pilot study (again translated to English) can be seen in \cref{fig:squad_par}.

\begin{figure}[ht]
    \noindent\rule{\linewidth}{0.5pt}
    \textbf{Paragraph with highlighted span:}
    
     All of Chopin's compositions include {\underline{the piano}}. Most are for solo piano, though he also wrote two piano concertos, a few chamber pieces, and some songs to Polish lyrics. 
    
    \textbf{Sample questions asked by our annotators:}
    
    What do all Chopin's songs include?
    
    What musical instrument will we hear in virtually all Chopin's compositions?
    
    \vspace{-0.2cm}\noindent\rule{\linewidth}{0.5pt}\vspace{-0.3cm}
    \caption{\label{fig:squad_par} Paragraph from SQuAD with two questions for
    the underlined span}
\end{figure}

We were mostly interested in spans of text which had more questions in SQuAD already because such spans seemed easier to create questions for. The distribution of questions per span in SQuAD can be seen in \cref{tab:squad_distribution}: the vast majority of spans has only one question and having more than four questions per span is very rare.
The rightmost column shows how many of such spans were included in our experimental data.

\begin{table}[ht]
    \centering
    \begin{tabular}{| c c c |}
        \hline
        Questions & Number of spans & Occurences in \\
        per span & in SQuAD 2.0 & experiment data \\
        \hline
        1&   81619& 15\\
        2&    2303& 15\\
        3&     166& 15\\
        4&      13& 10\\
        5&       8&  5\\
        6&       1&  0\\
        \hline
        Total: & 84110  & 60 \\
        \hline
    \end{tabular}
    \caption{\label{tab:squad_distribution}SQuAD 2.0 span distribution}
\end{table}

In total, 60 paragraphs were chosen from SQuAD 2.0 randomly but respecting the intended distribution in the third column in \cref{tab:squad_distribution}. This was to make sure that we included spans which had more than one corresponding questions. These 60 paragraphs were machine-translated to Czech and the spans were transferred to Czech manually. Bilingual users then had half of the SQuAD paragraphs in Czech and half in English, monolingual users saw only the Czech paragraphs. No user saw the same paragraph in both English and Czech.

\subsubsection{Annotation task composition}

The overall composition of types of stimuli is shown in \cref{tab:stimuli_composition}. The bilingual group received half of the SQuAD stimuli in Czech and half in English. The monolingual group received all the SQuAD stimuli in Czech.

All of the annotators overlap fully in technical and administrative issues. The monolingual annotators overlap fully within the group and 50\% with the bilingual group. Such overlaps are necessary for studying the same stimulus answers variations.

\begin{table}[ht]
    \centering
        \begin{tabular}{| l | c c |}
            \hline 
            Stimuli & monolingual & bilingual \\
            \hline
            Technical issues & 35 & 35 \\
            Administrative issues & 30 & 30 \\
            SQuAD 2.0 & 0 & 30 \\
            SQuAD 2.0 Czech & 60 & 30 \\
            \hline
            Total & 125 & 125 \\
            \hline
        \end{tabular}
    \caption{\label{tab:stimuli_composition} Overall composition of the input stimuli}
\end{table}

\section{Results}
\label{results}


Throughout the experiment, we recorded several types of data, while the users interacted with Ptakopět. The list of monitored events is in \cref{tab:annotation_info_1} and the description of each recorded information type is in \cref{tab:annotation_info_2}. Additionally, each logged event contained Unix timestamp.

\begin{table}[ht]
\small
    \begin{tabular}{|ll p{27mm} |}
        \hline
        Event code & Logged data & Description \\
        \hline
        START      & QUEUE & The user logs in \\
        NEXT       & SID, REASON & A stimulus is shown \\
        CONFIRM    & SID, TXT1, TXT2 & User accepts solution \\
        SKIP       & REASON & User skips stimulus \\
        TRANSLATE1 & TXT1, TXT2 & Forward translation is displayed \\
        TRANSLATE2 & TXT2, TXT3 & Backward translation is displayed \\
        ESTIMATE   & ESTIMATION & Quality estimation is highlighted \\
        ALIGN      & ALIGNMENT  & Source complexity is highlighted \\
        \hline
    \end{tabular}
    \caption{\label{tab:annotation_info_1} Logged information from Ptakopět users for each of their actions}
\end{table}
    
\begin{table}[ht]
    \bigskip 
    \begin{tabular}{| l p{54mm} |}
        \hline
        Logged data & Description \\
        \hline
        SID & Identifier of the relevant stimulus  \\
        TXT1 & Content of the source text area \\
        TXT2 & Content of the target text area \\
        TXT3 & Content of the backward translation text area \\
        ESTIMATION & Quality estimation data \\
        ALIGNMENT & Source to target word alignment \\
        REASON & User's motive for skipping answering the stimulus \\
        \hline
    \end{tabular}
    \caption{\label{tab:annotation_info_2} Description of logged information from Ptakopět users}
\end{table}

\vspace{0.0cm}

\subsection{Basic statistics}

We refer to sequences of log entries related to the same stimulus as segments. The number of finished segments, as well as their average duration in every domain, is shown in \cref{tab:segments_avg_duration}. Since the differences in duration between each segment was not high (min 90s, max 106s), we concluded that the users employed similar strategies across all domains and that no domain was exceptionally difficult nor easier than the others.

\begin{table}[t]
    \centering
    \begin{tabular}{| l c c |}
        \hline
        Domain & Segments & Average duration \\
        \hline
        SQuAD 2.0 & 141 & 100s  \\
        SQuAD 2.0 Czech & 346 & \ 94s \\
        Technical issues & 268 & 107s \\
        Administrative issues & 246 & \ 90s \\
        \hline
        All & 1001 & \ 98s \\
        \hline
    \end{tabular}
    \caption{\label{tab:segments_avg_duration} Number of segments and average duration per domain in collected data}
\end{table}

\subsection{Types of edits}

Some of the stimuli were skipped, mostly because the annotators did not have enough confidence in the MT system's performance (for a given stimulus) and were unable to produce a better result. We describe such segments as \textit{skipped} as opposed to \textit{finished}. From the finished ones, about a quarter of the segments were written linearly (no edits were performed, i.e. the annotator did not change or delete anything after linearly producing the input). Such segments are denoted as \textit{linear} as opposed to segments which had some edits in already written parts (\textit{with edits}). The number of skipped, finished, linear and edited segments can be seen in \cref{tab:dr1}.

We see that the proportion of skipped segments (i.e. segments where the annotator failed to produce an output they could accept) is not excessively high. The easiest to process were administrative issues (5.7\,\% skipped segments) and the hardest was the technical issues (10.8\,\%). SQuAD reached 7.8\,\% (English) and 7.5\,\% (Czech) of skipped segments.

Of the finished segments, most (72\%) were edited and not just linearly written (28\%). Additionally, in technical issues, the stimulus was the description of the technical problem itself, so the annotators could choose to simply copy this text and paste it in the input window. The number of occurrences of this behaviour is described in the table as \textit{init copy} (60\% of all edited). We also measured the number of final inputs which matched the initial stimulus (\textit{Copy \& submit}, 6\% of all edited).

\newcommand{\tbofall}{\multirow{2}{*}{\small (of all)}}
\newcommand{\tboffin}{\multirow{2}{*}{\small (of fin.)}}
\newcommand{\tbofedt}{\multirow{2}{*}{\small (of edt.)}}
\newcommand{\tbperc}[1]{\small (#1\%)}

\begin{table}[ht]
    \centering
    \begin{tabular}{| c l r r|}
        \hline
        Domain & Description & Segments & Ratio \\
    \hline
        \multirow{4}{*}{SQuAD 2.0} & Skipped & 11 \tbperc{8} & \tbofall{} \\
         & Finished & 130 \tbperc{92} & \\
         \cline{2-4}
         & Linear & 52 \tbperc{40} & \tboffin{} \\
         & With edits & 78 \tbperc{60} & \\
    \hline
        \multirow{4}{*}{\parbox{20mm}{\centering SQuAD 2.0 Czech}} & Skipped & 26 \tbperc{8} & \tbofall{} \\
         & Finished & 320 \tbperc{92} &\\
         \cline{2-4}
         & Linear & 110 \tbperc{34} & \tboffin{} \\
         & With edits & 210 \tbperc{66} & \\
    \hline
        \multirow{4}{*}{\parbox{20mm}{\centering Tech issues}} & Skipped & 29 \tbperc{11} & \tbofall{} \\
         & Finished & 239 \tbperc{89} & \\
         \cline{2-4}
         & Linear & 27 \tbperc{11} & \tboffin{} \\
         & With edits & 212 \tbperc{89} & \\
          \cline{2-4}
         & Init copy & 127 \tbperc{60} & \tbofedt{} \\
         & Copy \& submit \hspace{-0.3cm} & 13 \tbperc{6} & \\
    \hline
        \multirow{4}{*}{\parbox{20mm}{\centering Administrative issues}} & Skipped & 14 \tbperc{6} & \tbofall{} \\
         & Finished & 232 \tbperc{94} & \\
         \cline{2-4}
         & Linear & 70 \tbperc{30} & \tboffin{} \\
         & With edits & 162 \tbperc{70} & \\
    \hline
        \multirow{4}{*}{All} & Skipped & 80 \tbperc{8} & \tbofall{} \\
         & Finished & 921 \tbperc{92} & \\
         \cline{2-4}
         & Linear & 259 \tbperc{28} & \tboffin{} \\
         & With edits & 662 \tbperc{72} & \\
    \hline
    \end{tabular}
    \caption{\label{tab:dr1} Number of skipped, finished, linear and edited segments per domain in collected data together with percentage of all, finished or edited segments.}
\end{table}

We then focused on the finished segments which were later edited (not linear). We tried to extract the first input for which the annotator expected the translation to be successful, but edited later, because either the translation, backward translation or QE suggested that it may not be correct. We call this input \textit{first viable} and choose it heuristically as the longest nonfinal input ending with a punctuation mark. We then compute the similarity between the first viable source/translation and the final source/translation version as confirmed by the annotator using Gestalt Pattern Matching on word level (implemented in Python's difflib). This similarity is shown per domain in \cref{tab:similarity}.

\begin{table}[ht]
    \centering
    \begin{tabular}{| c c c |}
        \hline
        Domain & Input sim. & Translation sim. \\
    \hline
        SQuAD 2.0 & 69\% & 55\% \\
        SQuAD 2.0 Czech & 75\% & 60\% \\
        Tech issues & 78\% & 67\% \\
        Administrative issues & 74\% & 57\% \\
    \hline
        All & 75\% & 61\% \\
    \hline
    \end{tabular}
    \caption{\label{tab:similarity} Similarity between first viable and final versions of inputs and translations (only on segments with edits)}
\end{table}

From \cref{tab:similarity} we can see that even though the first viable and final inputs are quite similar (75\% on average across all domains), the first viable and final translations are less similar (61\% on average). Subject to general variance of sentences in Czech and German, this indicates that the edits had a considerable effect on the translation.

\subsection{Evaluation survey}

At the end of the experiment, we asked the annotators to fill in a short survey. The results are shown in \cref{tab:survey}.

\begin{table}[ht]
    \centering
    \begin{tabular}{| l | l | c |}
        \hline
        Question & Domain & Average \\
        \hline
        \multirow{3}{*}{\hspace{-0.1cm}\parbox{30mm}{\vspace{0.1cm}What confidence do you have in the translations you have created?}\hspace{-0.1cm}} & SQuAD 2.0 (both) & 1.14  \\
         & Technical issues & 2.86 \\
         & Administrative issues \hspace{-0.2cm} & 2.29 \\
         & All & 2.10 \\
         \hline \hline
         \multicolumn{2}{|l|}{\hspace{-0.1cm}\parbox{60mm}{\vspace{0.1cm}How useful was the highlighting of problematic words in technical issues?\\ \textbf{} \vspace{-0.2cm}}} & 2.29 \\
         \hline \hline
         \multicolumn{2}{|l|}{\hspace{-0.1cm}\parbox{60mm}{\vspace{0.1cm}Was the environment for these tasks useful, when compared to other web interfaces (Google Translate, Bing Translator and others)?}} & 1.71 \\
         \hline
    \end{tabular}
    \caption{\label{tab:survey} Annotator survey results (1 - most, 5 - least)}
\end{table}

We suspect that the overall results are affected by the relatively low quality of the MT system. Most of the annotators complained of this, stating that the MT system made obvious mistakes, such as adding random words. Should we deploy a better MT system, the average scores would probably go up. At the same time, it seems that we have chosen the right level of MT quality for the experiment: MT was not too good (edits were needed) and not too bad (at most 10.8\,\% of segments were given up).

The perceived confidence per domain confirms that technical issues were the hardest (probably because of vocabulary deficiency of the MT system in the IT domain) and it was the highest for encyclopedic questions.

Good news is that the overall usefulness of \ptakopet{} compared to standard web interfaces to MT was rated as 1.71 on the 1--5 (best--worst) scale, although the perceived utility of QE was lower (2.29).

We also inquired about the users' strategies. Most of them focused on the backtranslation to validate the output. If they suspected that the result might not be preferable (either by the backtranslation or by looking at the result itself), they tried reformulating the input by using synonyms. If that did not help, they tried simplifying the sentence, even beyond the threshold of a grammatically sound output sentence, attempting just to communicate the meaning properly.

It is worth noting that the backward translation can in principle fix previously introduced errors, thus hiding the problem. In these cases, the users could get a false sense of confidence in the translation. For such occasions, an external tool (e.g. MT quality estimator) is needed.

\subsection{Output validation}
\label{validation}

After we collected data from the previous annotation phase, we extracted final translations and translations of first viable inputs for each segment (if possible). We then asked another annotator with a good command of German (C2 on the CEFR scale) to rate each translation on the scale of 1 to 5 (best to worst) estimating to what extent the message would be understandable to Germans. 

\subsubsection{Validation results}

\begin{table}[ht]
    \centering
    \begin{tabular}{| c | c c | c c |}
    \hline
         & \multicolumn{2}{c|}{First viable} & \multicolumn{2}{c|}{Final} \\
    \hline
    Domain & Avg. & Var. & Avg. & Var. \\
    \hline
        SQuAD 2.0 & 3.43 & 2.56 & 1.91 & 2.00 \\
        SQuAD 2.0 Czech & 3.95 & 2.18 & 2.64 & 2.67 \\
        Tech issues & 3.77 & 1.79 & 3.10 & 2.23 \\
        Admin. issues & 4.05 & 1.91 & 2.92 & 2.55 \\
        \hline
        All & 3.85 & 2.07 & 2.77 & 2.55 \\
    \hline
    \end{tabular}
    \caption{\label{tab:qe_annotation_basic} Average quality ratings across domains for first viable and final translations  (1 - best, 5 - worst) }
\end{table}

The results for each domain for the final and first viable translations are in \cref{tab:qe_annotation_basic}. In each domain, the final translations were much better than the translations for first viable inputs. The average score improves from 3.85$\pm$1.44 to 2.77$\pm$1.60. Paired t-test showed that the difference is highly statistically significant (p < 0.0001 for 0.75 difference between final and first viable ratings).



\begin{figure}[ht]
    \centering
    \includegraphics[width=\linewidth]{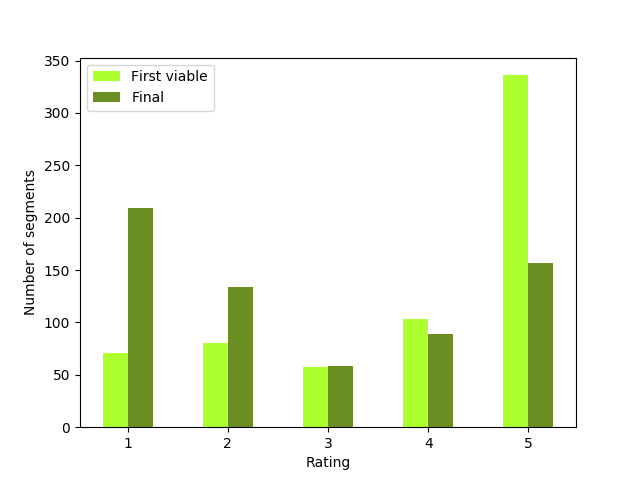}
    \caption{\label{fig:qa_buckets} Histogram of ratings for first viable and final translations (1 - best, 5 - worst)}
    \vspace{-0.2cm} 
\end{figure}

The validation scores assigned to the individual segments using a histogram is presented in \cref{fig:qa_buckets}. We see that the first viable translations received mostly the worst rating while final translations are bimodal: the majority received a favourable validation score but a considerable portion (24\%) had the worst score. We assume that in these cases, our setup was unreliable and fooled the user in accepting a misleading translation.

Overall, this is a clear success, as our technique helps people to produce better messages in a language they do not speak. Nevertheless, it is important to mention the limitations of our pilot study. Our heuristics for picking first viable inputs may include sentences which were actually not thought to be viable by the user. Maybe the sentences contained obvious errors, such as typos, which the user would fix anyway but maybe the user would not notice if we did not present the backtraslation. A more thorough exploration is needed to isolate such effects.

\subsubsection{Validation by sentence length}

One could expect that shorter sentences are generally easier to process by MT (except for very ambiguous very short sentences). To analyze this assumption in our setting, we plot the average validation score assigned to sentences based on the source length.

\begin{figure}[ht]
    \centering
    \includegraphics[width=\linewidth]{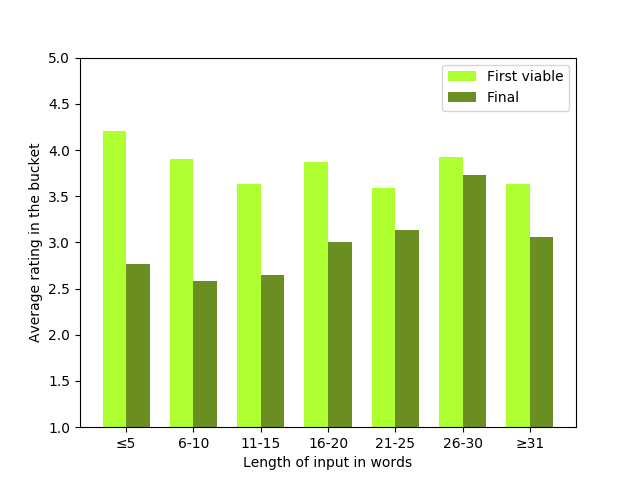}
    \caption{\label{fig:qa_sentence_length} Average rating for first viable and final translations based on the translated sentence length (1 - best, 5 - worst)}
    \vspace{-0.5cm}
\end{figure}

\cref{fig:qa_sentence_length} indicates that the assumed effect is not apparent in our case, at least not with our estimation of first viable inputs. The shorter sentences generally receive worse validation scores than longer ones, but the differences are not very big.

For final inputs, the assumption seems more accurate: The best validation score was assigned to sentences of 6--10 words and the worst to sentences over 25 words. A noteworthy observation is that for these long sentences, the improvement in the validation scores from first viable to final input is very low.


\vspace{-0.2cm} 

\section{Conclusion}
\label{conclusion}

In this paper, we presented Ptakopět, a modular system for outbound translation. \ptakopet{} allows users to produce messages in a language they do not speak and still gain some level of confidence in the resulting translation.

In a pilot experiment, users who did not know German were tasked to use this system for real-world use cases (communication with IT support, describing administrative issues and asking encyclopedic questions).

Across these domains, 5--10\,\% of inputs were problematic (our annotators have given up and skipped the stimulus). For the submitted translations, the average self-reported confidence in the translations was 2.10 on a 1--5 (best--worst) scale and the tool was found more useful than standard web interfaces to MT (average usefulness of 1.71, same scale).

The majority of inputs were edited and while initial inputs and the final inputs were quite similar in the source language (word-level Gestalt Pattern Matching similarity of 75\,\%), the translations of them differed more (average similarity of 61\,\%).

The second, validation, phase of our experiment confirmed that overall understandability of the translations improved from 3.9 to 2.71 on the 1--5 (best--worst) scale.





In future, we plan to refine the experiment design and also consider other features of the outbound translation user interface. For instance, we could directly estimate the chances of translating a word correctly by considering the number of occurrences in the training corpus of the underlying MT systems, or we could offer synonyms to poorly covered source words (based on a sizeable monolingual corpus). The evaluation could also contrast how much each of these features helps in the task of producing a message in an unknown language.





\section*{Acknowledgments}
This project has received funding from the grants
H2020-ICT-2018-2-825303 (Bergamot) of the European Union and
19-26934X (NEUREM3) of the
Czech Science
Foundation.

We used language resources developed and/or stored and/or distributed by the LINDAT-Clarin project of the Ministry of Education of the Czech Republic (project LM2010013).

\vspace{0.8cm}

\section*{Bibliographical References}

\bibliographystyle{lrec}
\bibliography{biblio}

\end{document}